# Understanding the temporal evolution of COVID-19 research through machine learning and natural language processing

Ashkan Ebadi[1,4,*], Pengcheng Xi[2], Stéphane Tremblay[2], Bruce Spencer[3,5], Raman Pall[2], and Alexander Wong[6,7]

[1] National Research Council Canada, Montreal, QC H3T 1J4, Canada
[2] National Research Council Canada, Ottawa, ON K1K 2E1, Canada
[3] National Research Council Canada, Fredericton, NB E3B 9W4, Canada
[4] Concordia University, Concordia Institute for Information Systems Engineering, Montreal, QC H3G 2W1 Canada
[5] University of New Brunswick, Faculty of Computer Science, Fredericton, NB E3B 5A3, Canada
[6] University of Waterloo, Department of Systems Design Engineering, Waterloo, ON N2L 3G1, Canada
[7] Waterloo Artificial Intelligence Institute, Waterloo, ON N2L 3G1, Canada
[*] Email: ashkan.ebadi@nrc-cnrc.gc.ca

**Abstract**
The outbreak of the novel coronavirus disease 2019 (COVID-19), caused by the severe acute respiratory syndrome coronavirus 2 (SARS-CoV-2) has been continuously affecting human lives and communities around the world in many ways, from cities under lockdown to new social experiences. Although in most cases COVID-19 results in mild illness, it has drawn global attention due to the extremely contagious nature of SARS-CoV-2. Governments and healthcare professionals, along with people and society as a whole, have taken any measures to break the chain of transition and flatten the epidemic curve. In this study, we used multiple data sources, i.e., PubMed and ArXiv, and built several machine learning models to characterize the landscape of current COVID-19 research by identifying the latent topics and analyzing the temporal evolution of the extracted research themes, publications similarity, and sentiments, within the time-frame of January- May 2020. Our findings confirm the types of research available in PubMed and ArXiv differ significantly, with the former exhibiting greater diversity in terms of COVID-19 related issues and the latter focusing more on intelligent systems/tools to predict/diagnose COVID-19. The special attention of the research community to the high-risk groups and people with complications was also confirmed.

**Keywords**
COVID-19 research landscape, Topics evolution, Machine learning, Structural topic modeling, Text mining

**Introduction**
The ongoing pandemic of the coronavirus disease 2019 (COVID-19), caused by the severe acute respiratory syndrome coronavirus 2 (SARS-CoV-2), has been affecting human lives and communities around the world, causing global social and economic disruption (International Monetary Fund, 2020). The first case of COVID-19 can be traced back to Wuhan (China) in December 2019 (Hui et al., 2020). The World Health Organization (WHO) declared the outbreak in January 2020 and characterized as a pandemic in March 2020 (World Health Organization, 2020). As of June 2020, more than 6.5 million COVID-19 cases have been reported worldwide resulting in more than 500,000 deaths, as of this writing (Johns Hopkins University, 2020), with numbers increasing daily.

The first COVID-19 case in Canada was identified in January 2020 (Government of Canada, 2020). Although most of the COVID-19 positive Canadian cases are in the most populous provinces with Quebec and Ontario being the top-two, as of June 2020 there have been confirmed cases in all the Canadian provinces and territories except for Nunavut (Government of Canada, 2020). With cases of community transmission being confirmed, all Canadian provinces and territories have declared states of emergency or public health emergency in one form or another in March 2020 (National Post, 2020).

Governments and authorities worldwide are actively fighting against the disease by implementing various measures and policies such as travel restrictions and facility closures. The scientific community has also responded to the COVID-19 pandemic in different ways. Although medical science, drug discovery, and epidemiology have seen the most attention, the COVID-19 pandemic is a multidimensional phenomenon and as such has strong socio-economic, psycho-social, and technological implications (Zhang & Shaw, 2020). Examples of socialeconomic issues caused by the COVID-19 pandemic range from cancellation of sports, political, and cultural events to city lockdowns and supply shortages due to panic buying (Council on Foreign Relations, 2020; Yuen et al., 2020). Apart from setting policies, governments are investing significantly on COVID-19 research. In Canada alone, the federal government is investing $1.1 billion in research and development on COVID-19 and vaccine development (CityNews, 2020).

The healthcare and medical research communities have rapidly and widely responded to the COVID-19 challenge since the beginning. Drug and vaccine discovery research (Dong et al., 2020; Gao et al., 2020), analyzing the impact of the disease on people who are suffering from other diseases or complications (Fang et al., 2020; Klok et al., 2020), as well as on high-risk groups such as older persons (Applegate & Ouslander, 2020), are but a few examples of the comprehensive effort of the medical community towards fighting against the disease. Additionally, researchers are also investigating the effects of the COVID-19 outbreak on people's mental health state (Huang & Zhao, 2020), including the medical staff (Chen et al., 2020), who are in the front line of the fight against the disease.

Today, artificial intelligence (AI) and emerging technologies play a key role in responding to the COVID-19 crisis from accelerating drug research to diagnosing the disease. The AI systems have used both handcrafted features and deep learning features for screening patients and severity assessment. In a study on distinguishing COVID-19 patients from community-acquired pneumonia (CAP) (Shi, Xia, et al., 2020), the authors extracted a list of handcrafted features from computed tomography (CT) scans, including volume, features, histogram and surface features. In another study on assessing severity (Tang et al., 2020), chest CT scans were first segmented, and then quantitative metrics were calculated, including infection volumes and percentage of infection (POIs). Wang and Wong (2020) designed a tailored deep convolutional neural network named "COVID-Net" for distinguishing normal, pneumonia, and COVID-19 patients using chest X-ray images. Further to that, Wong et al. (2020) extended the use of COVID-Net for severity assessment through training and validating the network for geographic extent and opacity extent scoring of chest X-rays.

Motivated to gain a better understanding about the current COVID-19 research landscape, this study leveraged natural language processing (NLP) and machine learning to analyze the evolution of COVID-19 research in a quantitative manner based on scientific publications from two key data sources, i.e., PubMed and ArXiv. To the best of our knowledge, this is the first study that considers multiple data sources of different nature to characterize the landscape of COVID-19 research and to investigate its evolution over time, including but not limited to the similarity between the performed research over weekly time intervals as well as the emotional trajectory. Our approach has the potential to provide key stake-holders and decision-makers with a clear mapping of the COVID-19 research landscape and identify the main COVID-19 research themes, their temporal progression, evolution, and novelty. It also sheds light on the emotional dimension hidden in the performed research. This would assist the policy-makers to direct and adjust their strategies if required. The remainder of this study is as follows. First, the "Data and methodology" section describes the data and techniques in more detail. Second, the "Results" section presents findings of the research. Third, we discuss our findings and present our conclusions in the "Discussion and conclusion" section. Fourth and finally, we present some future directions and limitation of the research in the "Limitations and future work" section.

**Data and methodology**
The scope of this study covers all COVID-19 related publications accessible through ArXiv[1] and PubMed[2] services. The metholodgy has four main steps, which are discussed in detail below.

**Data collection and filtration**
We initially collected all the articles published in 2019 and 2020 from the aforementioned sources and dropped articles with no/incomplete publication dates as well as those with neither titles nor abstracts. Next, we removed duplicated articles and only included publications in 2020 due to the data sparsity in 2019. This resulted in a total of 14,172 publications as of May 31, 2020.

**Text pre-processing**
We merged the titles and abstracts of the collected publications and applied several pre-processing steps, e.g., converting the text to lowercase, correcting special characters, removing stop words using a customized English stop words list, and punctuations. We decided to use both titles and abstracts for the analyses as although the abstract is a condensed representation of the articles and contains more information, the title may also contain some informative keywords/keyphrases that are not present in the abstract. As such, integrating both titles and abstracts provides us with more information to build a better understanding of the COVID-19 research landscape. The processed textual data were tokenized, and a document-term frequency matrix was generated.

**Descriptive and temporal text analyses**
We then performed descriptive analyses on the collected data investigating publication trends in the examined data sources as well as extracting target countries mentioned in the publications. Next, we did temporal text analyses to investigate keyphrases patterns, publication sentiments, and research similarities over time. We used the TextBlob Python package (Loria, 2018) to extract sentiment from the publications. The sentiment score is within the [-1.0, 1.0] range, where 0 indicates neutral, -1 indicates very negative sentiment, and +1 indicates very positive sentiment. We trained a Doc2Vec model (Le & Mikolov, 2014) on the corpus to learn publication-level embeddings and assess the similarity between publications by calculating the cosine similarity between the embedding vectors. We aggregated the sentiment analyses and publication similarity results by week and analyzed their trends.

**Structural topic modeling**
After descriptive and temporal text analyses, we did topic modeling to extract the main research themes. As an unsupervised machine learning technique, topic modeling can find latent semantic topics in huge text data collections, summarize the corpus automatically, and extract knowledge (Blei et al., 2003). We used structural topic modeling (STM) to extract topics as it lets topics to be correlated and it also allows us to incorporate document-level covariates of interest (Roberts et al., 2014), e.g., publication date in our case. Such properties were critical for our research objectives as they enabled us to capture the hidden temporal aspect necessary to analyze research topics evolution. We built three STM models: 1) STM model built on the entire dataset with a monthly granularity, 2) STM model built only on PubMed dataset with a weekly granularity, and 3) STM model built on ArXiv dataset with a weekly granularity. We will refer to these models as STM-ALL, STM-PUBMED, and STM-ARXIV in the rest of this study, respectively. Findings from the three models were complementary, each contributing to a better mapping of the state of research in the target publications.

---
[1] ArXiv is an open-access archive for scholarly articles. For more information, please see: https://arxiv.org/
[2] PubMed is a free search engine over the MEDLINE database of references on life sciences topics. For more information, please refer to: https://pubmed.ncbi.nlm.nih.gov/

The number of topics needs to be set in advance in STM as a fixed parameter. There is no general consensus in setting the optimal number of topics in topic models (Lucas et al., 2015); however, the choice is highly dependent on the application and objectives. A completely automatic approach to find the optimal number of topics might not be very accurate (Maskeri et al., 2008) and it often needs human intervention. We followed a multi-layer approach to determine the number of topics, similar to the one proposed in Ebadi et al. (2020). We first built several baseline latent Dirichlet allocation (LDA) models (Blei et al., 2003) with a different number of topics in the range of [2, 10] and used an intrinsic evaluation metric, i.e. topic coherence, to quantitatively evaluate them. In particular, we used the $C_v$ coherence score proposed in Röder et al. (2015). We did this analyses for each of the three data scenarios, i.e. the entire data, PubMed only, and ArXiv only.

For the entire data scenario, the coherence measure peaked at 3 and 7. We decided not to consider 3 as the optimal number of topics since it provided very generic topics. As the next step, we considered the vicinity of 7, i.e. [6, 8] and followed Roberts et al. (2014) approach running several automated tests using multiple criteria such as the exclusivity of the topics, to refine the range further. This narrowed down the range for the optimal number of topics to [7, 8]. Three domain experts were then provided with the results, verified the models, checked keywords, and keyphrases assigned to each topic, analyzed topic-document distributions, and assessed the quality of the models. Based on their assessment, the optimal number of topics for the entire corpus was found to be 7, therefore, we used the STM model with 7 topics as the final model (STM-ALL) for the entire data scenario. The same analyses for the other two data scenarios, i.e. PubMed only and ArXiv only, resulted in 7 and 4 topics, respectively. The three experts manually labeled the generated topics after careful examination of the extensive set of keywords for each topic. We used more than one expert as well as an odd number of experts to reduce the subjectivity effect in labeling topics as well as setting the optimal number of topics. Using the extracted topics, we finally analyzed topic evolutions over time. The conceptual flow of the study is depicted in Figure 1.

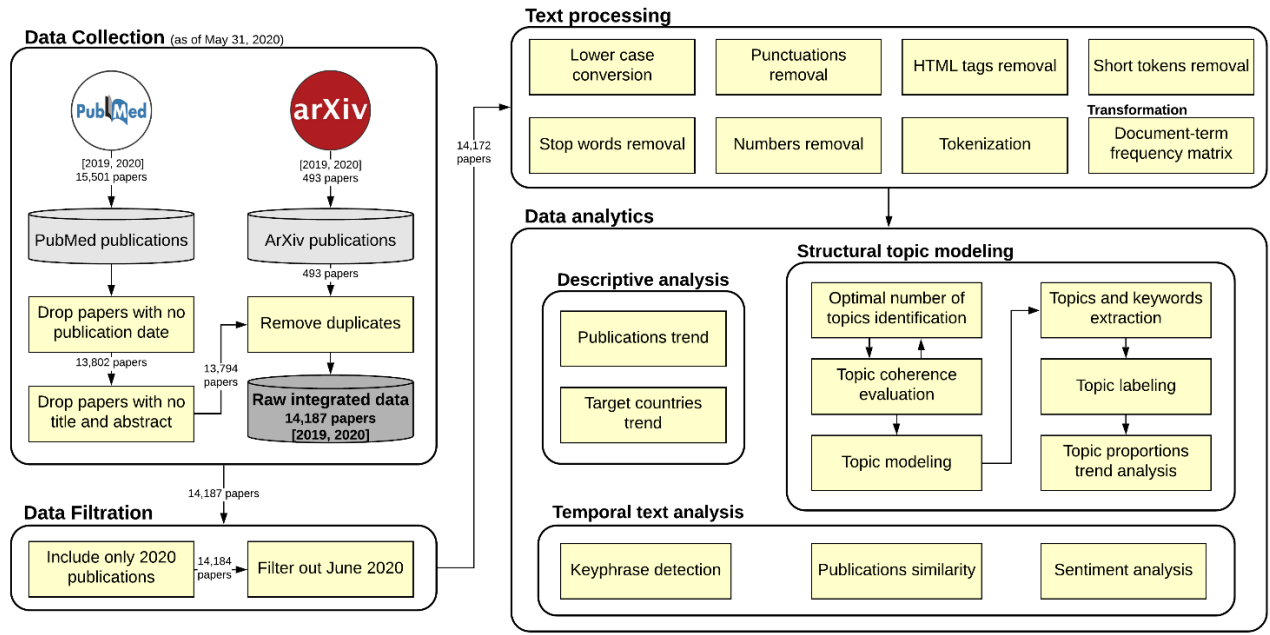

**Figure 1.** The analytical flow. The pipeline contains four main steps, i.e. data collection, data filtration, text processing, and data analytics. In the data collection step, COVID-19 publications within the period of 2019–2020 are collected. Data is then filtered to only contain publications in 2020 and data is passed to the data analytics step. After performing descriptive analyses, we did temporal text analyses to identify keyphrases and assess publications' similarity and sentiment over time. The optimal number of topics is determined and set in the structural topic modeling component where the final STM models are built to extract the main research themes, their keyword sets, as well as the temporal trends.

## Results

### Descriptive analyses
Our target dataset contained 14,172 scientific publications about COVID-19. In this section, we present the results of the descriptive analyses.

*COVID-19 publications trend*
Figure 2 shows the publication distribution separately for PubMed (the orange solid line) and ArXiv (the blue dashed line) over months. The bold numbers on the lines show the number of publications for the respective month. As seen, the number of publications follows a sharp increasing trend, except for ArXiv in the final period where a slight decrease is observed. There is only one article published in January 2020 in our final dataset. The drastic increase might reflect several time-related aspects such as the importance of the issue, the interest of the scientific community, and the dimensions of the problem that have augmented over time.

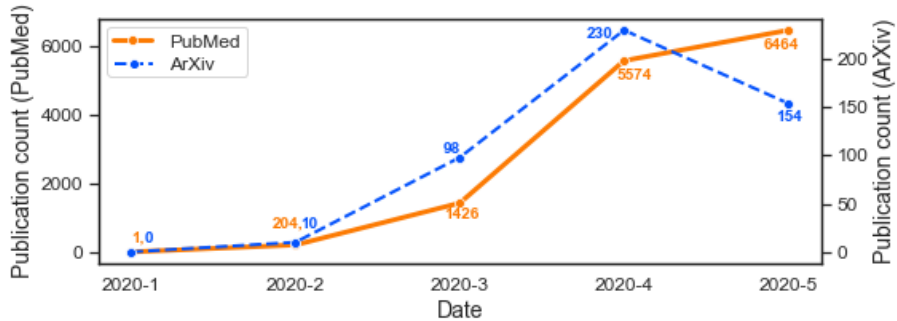

**Figure 2.** The trend of COVID-19 publications, from January to May 2020, in ArXiv (blue dashed line) and PubMed (orange solid line). The figure has two y-axes, the left y-axis represents PubMed and the right y-axis represents ArXiv data.

*Geographic distribution*
Using NLP techniques, we extracted country names that were mentioned in the titles and abstracts of the publications to analyze the geographic distribution of the target countries. As seen in Figure 3-a, China has been mentioned significantly higher than the other countries in the papers over the entire examined period. Italy and the United States rank 2nd and 3rd, respectively. From Figure 3-a, it can be observed that the scientific community responded rapidly to the pandemic from a research front, focusing on various countries' data. China is mentioned most frequently since it has many cases and is the site of the original outbreak, while Italy, the United States, India, and other countries are also mentioned frequently for their outbreaks and active cases. To further investigate, we focused on the top-6 countries in Figure 3-b and analyzed the temporal trend of the geographic distribution in those countries. With the increasing trend of publications within February-April 2020 (Figure 2), the scientific community's focus on the top-6 countries has also increased Figure 3-b). However, a constant or decreasing trend is observed after April 2020. This is in line with our previous observation that the researchers have dynamically responded to the pandemic over time based on its geographic movement. The decreasing trend may also explain the fact that the pandemic has become global so that specific countries' outbreaks are less often the focus of the COVID-19 research papers.

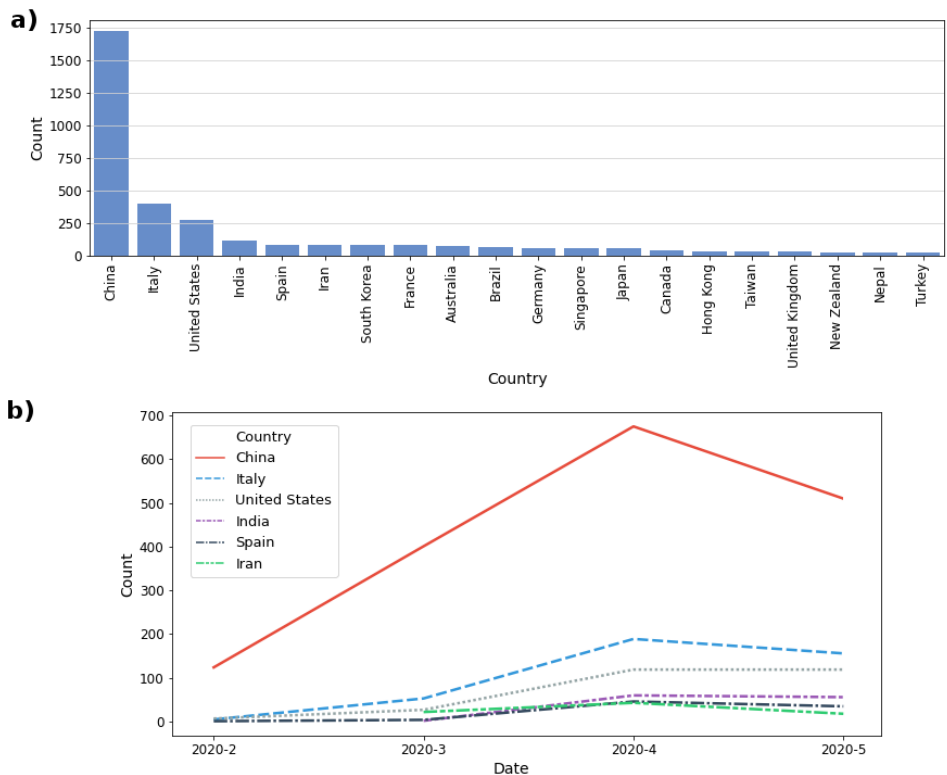

**Figure 3. a)** Distribution of the countries mentioned in the titles and abstracts of the publications, and **b)** Temporal trend of the top-6 most mentioned countries in the publications.

*Keyphrases over time*

As a preliminary step in investigating the vocabulary evolution in COVID-19 related publications, we identified and examined keyphrases that are present in the publications. We extracted n-grams of length 2 to 4, a maximum of 200, for each month from January to May 2020, and sorted the keyphrases out based on their frequency and performed percentage normalization. We filtered out keyphrases that contained highly frequent keywords such as "covid" and "coronavirus" as they were not informative for this analyses. We dropped January from the analyses as only six keyphrases were extracted due to the limited number of publications. Figure 4 shows the results. The numbers on the bars in the figure reflect the exact frequency of the respective keyphrase. Several initial observations are made: 1) While in the beginning scientific community seems to focus more on the pandemic aspect of the disease and its acute, imminent danger to public health, over time the attention has been gradually drawn to longer-term and chronic impacts on the public, such as mental health, 2) Different regions and countries that are seen among the keyphrases in different time intervals are correlated with the prevalence of the disease and number of confirmed cases in those regions, 3) In the final period, apart from clinical trials that might be due to COVID-19 vaccine generation attempts, the impact of different policies such as social distancing has attracted the researchers' attention. Of course, these findings are preliminary and we will further investigate them in the next section.

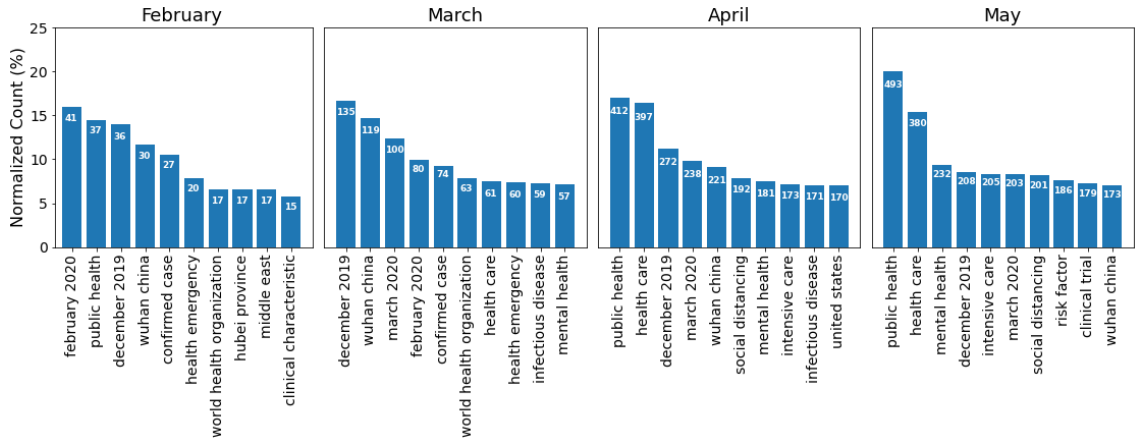

**Figure 4.** The top-10 most frequent keyphrases in COVID-19 publications. The numbers on the bars represent the exact frequency of the respective keyphrase.

**Research similarity, sentiment, and topics evolution**

*The entire dataset, PubMed and ArXiv, monthly granularity*

We employed structural topic modeling (STM) technique (Roberts et al., 2014) and extracted seven research topics, using the month of publication as the covariate (STM-ALL model). To improve the quality of the keywords, we followed Bischof & Airoldi (2012) approach and extracted keywords that were not only frequent but exclusive as well. As explained in the "Data and methodology" section, the extracted topics were verified by three domain experts, and a representative label was generated for each topic. The seven labeled topics are: 1) Oncology, 2) Personal protective equipment (PPE), 3) Analytics, 4) Rehabilitation-panic, 5) High-risk groups, 6) Genomics, and 7) Intubation-oxygenation. One may note that these research topics only represent the main areas of interest of the researchers at an abstract level and in no means, they capture all the details about the performed research.

Having the STM model built and the topics extracted, we regressed the proportion of each publication on the date of publication, i.e., the publication-specific covariate to analyze the evolution of topics over time. In other words, we estimated the conditional expectation of topic prevalence given the characteristics of the publication and date of publication. The results are depicted in Figure 5-a. The shaded areas between the dotted lines in the figure represent the 95% confidence interval. As observed, three topics, i.e., intubation-oxygenation, analytics, and rehabilitation-panic, followed a decreasing trend over time while the others' prevalence increased. Analytics and rehabilitation-panic decreased more slightly than intubation-oxygenation. In the beginning, intubation-oxygenation was the main focus in researchers' publications, however, in the final period, more attention was drawn to the high-risk groups. Moreover, researchers have almost constantly focused on genomics within the examined time interval, placing it among the top-3 most prevalent topics in all periods. We also investigated the distribution of dominant topics across publications over time to complement the previous findings. Each publication can belong to more than one topic in structural topic modeling, and topics are assigned to each publication with a probability. We extracted the publication-topic probability matrix from the generated topic model and for each publication, we assigned the topic with the highest probability to it. Figure 5-b shows the distribution of the dominant topics. Genomics was the only topic observed in January due to data sparsity. Overall, it can be observed that oncology, high-risk group studies, and genomics have been the most dominating topics. Researchers have constantly considered genomics as one of the main areas of research with regards to COVID-19. Over time, oncology, personal protective equipment, and studying the high-risk groups have attracted more attention. Although in the beginning intubation-oxygenation was one of the main research areas, researchers focused more on other areas along time. Despite some fluctuations, an almost steady trend is observed for analytics after February 2020.

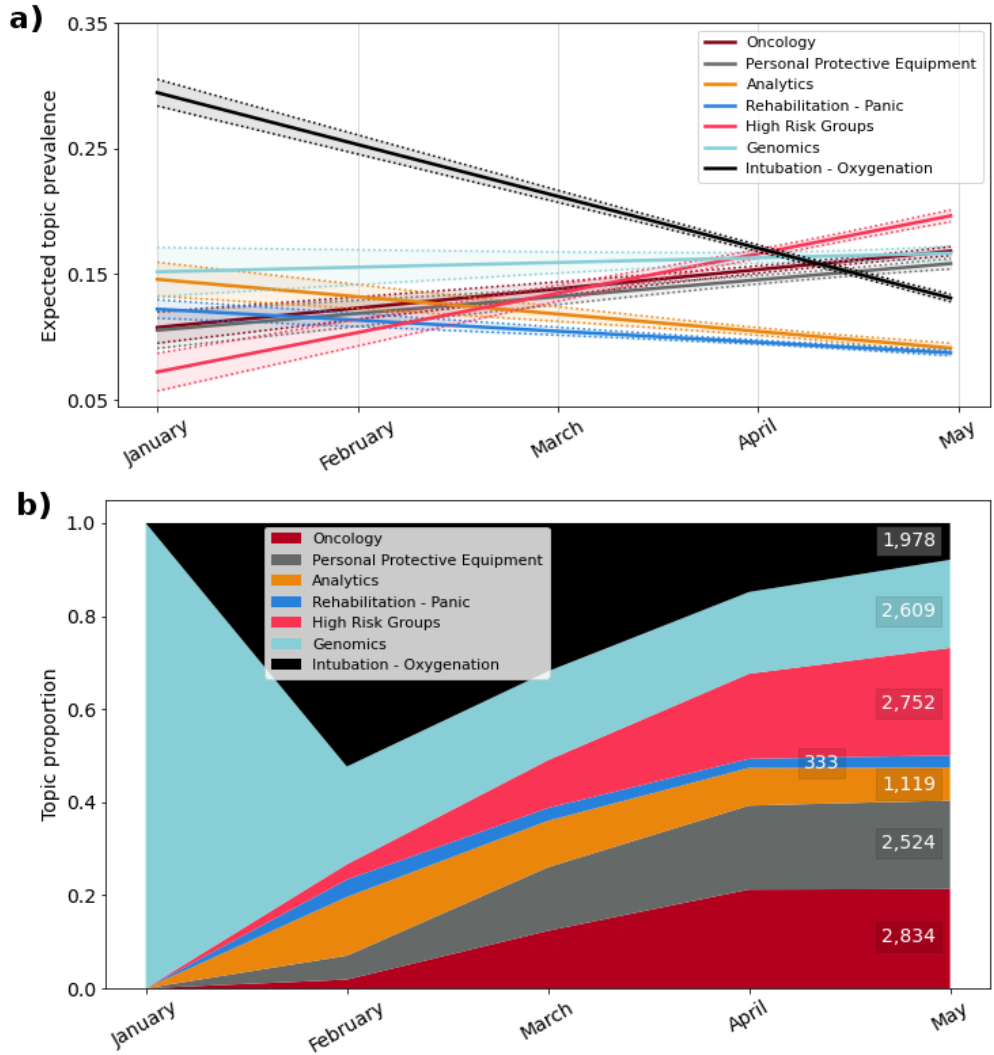

**Figure 5. a)** Topic prevalence in COVID-19 publications from January to May 2020, the STM-ALL model. The shaded areas between dotted lines indicate the 95% confidence interval, and **b)** Dominant topic distribution across publications over time, the STM-ALL model. The numbers on the figure represent the total number of publications dominated by the respective topic.

*PubMed only, weekly granularity*

In this section, we only focus on the PubMed publications and present the results of the STM-PUBMED topic model as well as sentiment and research similarity trends over the consecutive weeks in the period of January-May 2020. To build the topic model, we followed the same process as discussed in the previous section, except for the granularity level that is weekly here, and extracted seven main topics from the PubMed corpus as follows: 1) Panic pandemic, 2) Social services and emergency, 3) Genomics – drugs, 4) High-risk groups, 5) Rehabilitation, 6) Pregnant women – hospitalization, and 7) Surgical care. Figure 6-a shows the estimated conditional expectations of topics prevalence every week. The shaded areas between the dotted lines in the figure represent the 95% confidence interval. As seen, only two topics, i.e. surgical care and social services and emergency, have followed an increasing trend over time while the other topics' prevalence has declined. The range of the expected topic prevalence is not very wide, i.e. in [0.08, 0.21]. From the analyses, it can be observed that at the beginning, more attention was focused on the high-risk groups; however, in the final period, the focus shifted more towards surgical care.

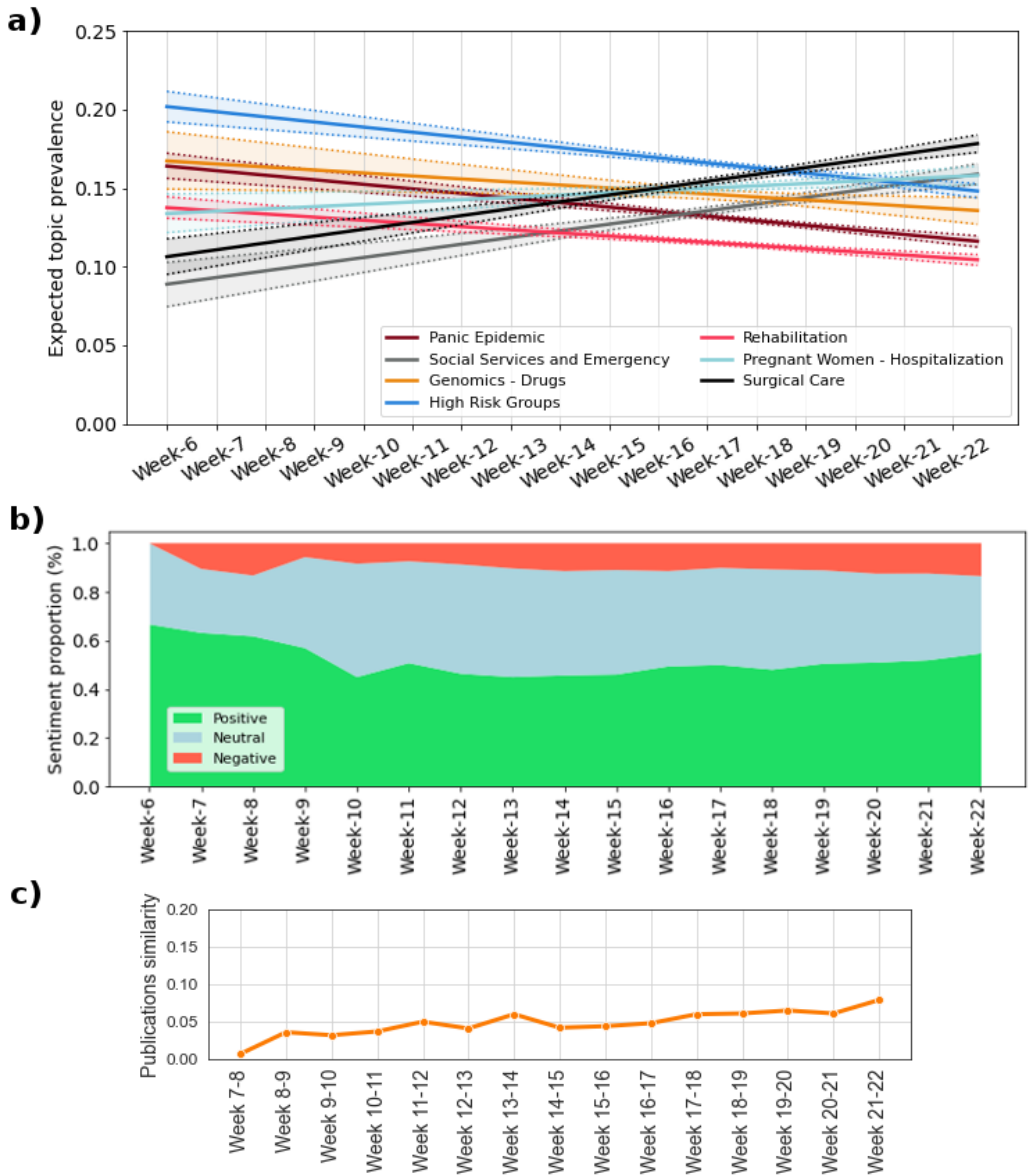

**Figure 6. a**) Topic prevalence in COVID-19 publications, the STM-PUBMED model, weekly granularity. The shaded areas between dotted lines indicate the 95% confidence interval, **b**) Sentiment percentages in PubMed publications, and **c**) Publications similarity in PubMed dataset.

Figure 6-b shows the sentiment percentages in PubMed publications over the examined period. In Week-6, ~65% of the detected sentiment was positive having the rest as neutral. However, along time some negative sentiment is detected such that at the final period (Week-22), there is ~10% negative, ~30% neural, and ~60% positive sentiment in the PubMed publications. An almost steady trend is observed for the negative sentiment after Week-9. Although the positive sentiment declined from ~65% to ~45% from Week-6 to Week-10, it has maintained a level of ~50-60% afterward. Publication similarity in the PubMed dataset is shown in Figure 6-c. Although the similarity between the publications increases over time, in general, the similarity score is not high as it was lower than 0.1 in all the periods. This may indicate a higher variety and a wider scope of the research that is published in PubMed. That is, the content overlap between PubMed publications (considering only the title and abstract) is not very high. Additionally, the figure may also indicate a higher specificity of the research published in PubMed.

*ArXiv only, weekly granularity*
In this section, we present the results of the STM-ARXIV topic model as well as sentiment and research similarity trends over the consecutive weeks in the period of January-May 2020. We followed the same process as discussed in the previous section, i.e. with a weekly granularity. Four main topics were extracted from the ArXiv corpus: 1) Contagion projection, 2) Deep learning – medical imaging, 3) Drugs, 4) Social media – misinformation. The estimated conditional expectations of topics prevalence are depicted in Figure 7-a. The shaded areas between the dotted lines in the figure represent the 95% confidence interval. As seen, the proportion of the social media – misinformation topic has increased over time. A slightly increasing trend is also observed for the deep learning – medical imaging topic. Although the contagion projection was the most prevalent topic, in the beginning, it ranked the 3rd in the final period. Compared to Figure 6-a, wider regions of confidence intervals are observed in Figure 7-a that could be due to the smaller dataset. Also, the range of the topic prevalence is wider in Figure 7-a compared to Figure 6-a.

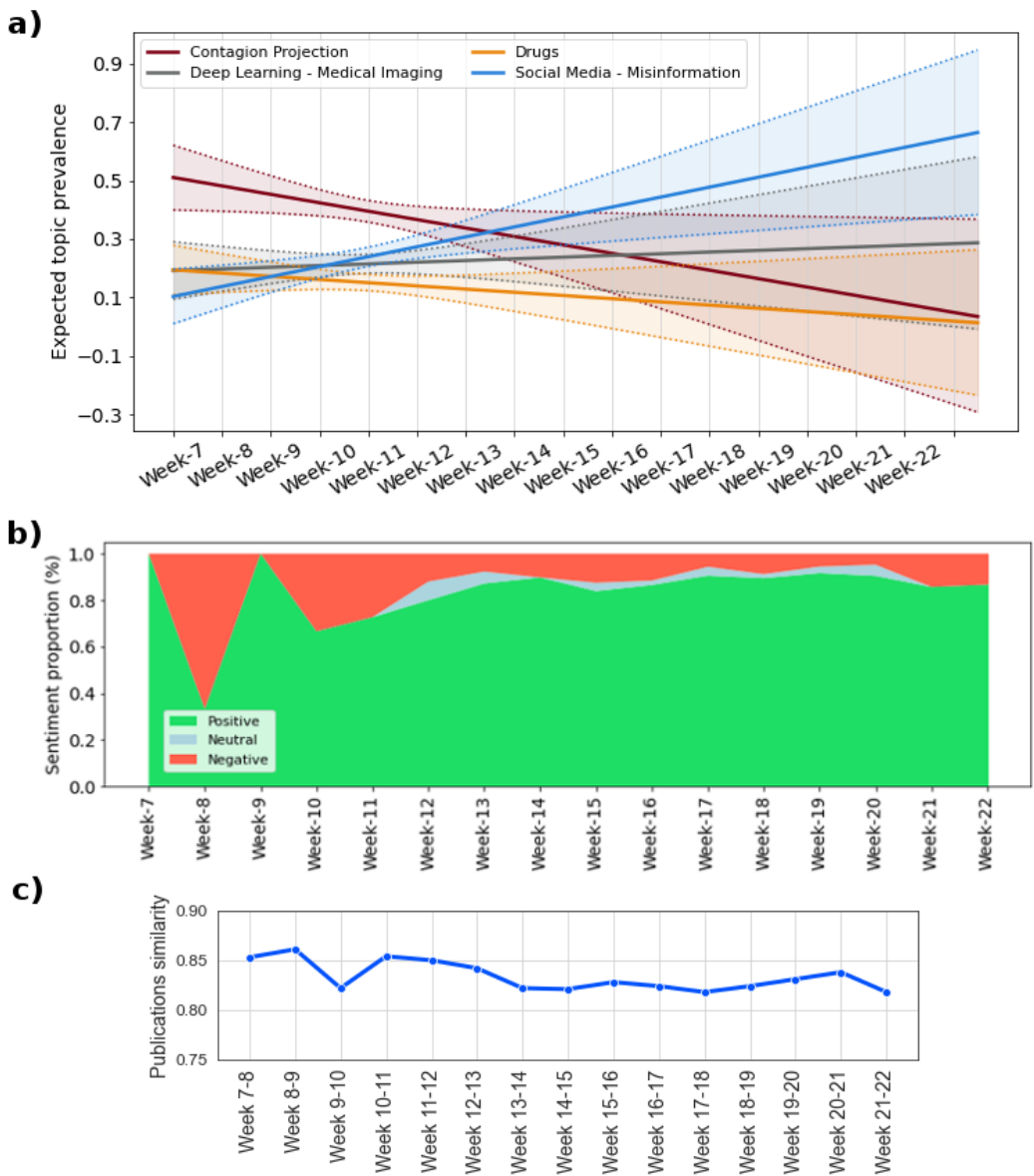

**Figure 7. a)** Topic prevalence in COVID-19 publications, the STM-ARXIV model, weekly granularity. The shaded areas between dotted lines indicate the 95% confidence interval, **b)** Sentiment percentages in ArXiv publications, and **c)** Publications similarity in the ArXiv dataset.

The sentiment percentages in ArXiv publications are shown in Figure 7-b. Overall, the percentage of neutral sentiment is negligible compared to positive and negative sentiments. More sentiment fluctuation is seen at the beginning of the examined period, having ~65% of negative sentiment in the Week-8. After the fluctuations, the proportion of the positive sentiment increased and reached ~80% in the final period. Comparing Figure 7-b with Figure 6-b, a higher positive sentiment ratio is observed in the ArXiv dataset. This may indicate the different nature of PubMed and ArXiv data sources. Such difference might be due to the fact that COVID-19 publications in ArXiv seem to be more oriented toward analytics and computer science (Figure 7-a). Figure 7-c shows publications similarity in the ArXiv dataset. The similarity between publications slightly decreases over time, however, the level of similarity is relatively high in all periods, being in the range of [0.8, 0.9]. The high similarity partially confirms our previous findings as the ArXiv publications seem to be more oriented towards statistical and computer science algorithms and fine-tuning them for COVID-19.

**Discussion and conclusion**

As the COVID-19 crisis continues, the research community is actively responding aiming to contribute to the wellness of the society as well as patients' outcome. The volume of research publications related to COVID-19 produced only in the first months of 2020 has been tremendous and has targeted a diverse set of issues. In this study, we focused on two different data sources, i.e. PubMed and ArXiv, and employed machine learning and natural language processing techniques to better understand the landscape of COVID-19 research and its evolution over time. Our comprehensive analyses, performed at different levels of granularity, could assist the decision- and policymakers to better understand COVID-19 research dynamics that might help to set or adjust strategies.

Considering all the extracted topics from our different models, it is clear that the research community has continuously focused on the vulnerable and high-risk populations who are in danger of severe illness from COVID-19. This was reflected by multiple topics in our models such as high-risk groups, pregnant women, and surgical care. Due to the importance of the matter, we suggest continuous monitoring of the performed research to ensure the wellness of the groups at particular risk from COVID-19, e.g. older people, pregnant women, patients with medical complications, and make sure that they are not left behind in the COVID-19 response.

Our findings confirmed that different types of research are being published in PubMed and ArXiv. While the latter, as an open-access repository with very fast processing time, hosts more technical papers that aim to detect/diagnose COVID-19 or predict its spread, PubMed was found to be hosting a diverse set of medical papers targeting a wider set of issues related to COVID-19. Low similarity observed among PubMed publications might be an indication of the diversity of research as well as the fast rate of change in the topics/issues that the medical community is focusing on regarding the COVID-19 crisis. Different sentiment patterns also confirm differences between types of research published in PubMed and ArXiv. Therefore, considering multiple data sources in similar research could be beneficial as the findings could be complementary.

The high similarity that was observed among ArXiv publications over time along with the extracted topics highlights the importance of advanced analytics and deep learning techniques and their application to COVID-19 medical images. In screening COVID-19 patients, medical imaging plays a complementary role to Reverse Transcription-Polymerase Chain Reaction (RT-PCR), the gold standard of confirming COVID-19 patients. In particular, we see increasing use of artificial intelligence (AI) in medical imaging for improving the efficiency of radiologists and for increased accuracy in diagnosis. Shi et al. (2020) conducted a review on AI techniques in imaging data acquisition, segmentation, and diagnosis for COVID-19. In image acquisition, AI can be used to automate the scanning procedure and avoid physical contact between patients and radiologists. AI can also improve efficiency through the accurate delineation of infections in X-ray and CT images. Subsequent analyses of the segmented abnormalities can help radiologists make clinical decisions for disease tracking and prognosis.

Lastly, the COVID-19 crisis has not only attracted the attention of the scientific community but the public's as well. Despite the advantages of such global attention, this has resulted in the spread of misinformation on social media. Even though it is often unintentional, making/spreading bad information could cause severe harm to society. Interestingly, researchers have been also working in this area providing tools to distinguish between good and bad information, as reflected by the social media-misinformation topic in the ArXiv dataset.

**Limitations and future work**

We included articles published in January-May 2020. Similar research could be performed in various snapshots in a year as more data become available. The findings of this research may only reflect the researchers' focus to COVID-19 at a very high level. Other levels of abstraction could be considered in future research using our proposed methodology. Another future direction would be country-specific analyses. Also, as more data become available several other variables of interest could be included in the analyses to examine different scenarios such as the impact of government policies over time. Research topics fusion and/or division along time can be evaluated in the future as well. Future research may consider the full body of the articles to perform the analyses. Analyzing the methods section, in particular, might be informative revealing methodological evolution.

Wang, L., & Wong, A. (2020). COVID-Net: A Tailored Deep Convolutional Neural Network Design for Detection of COVID-19 Cases from Chest X-Ray Images. *ArXiv:2003.09871 [Cs, Eess]*. http://arxiv.org/abs/2003.09871

Wong, A., Lin, Z. Q., Wang, L., Chung, A. G., Shen, B., Abbasi, A., Hoshmand-Kochi, M., & Duong, T. Q. (2020). Towards computer-aided severity assessment: Training and validation of deep neural networks for geographic extent and opacity extent scoring of chest X-rays for SARS-CoV-2 lung disease severity. *ArXiv:2005.12855 [Cs, Eess]*. http://arxiv.org/abs/2005.12855

World Health Organization. (2020). *Statement of the Twenty-Fourth IHR Emergency Committee*. https://www.who.int/news-room/detail/08-04-2020-statement-of-the-twenty-fourth-ihr-emergency-committee

Yuen, K. F., Wang, X., Ma, F., & Li, K. X. (2020). The Psychological Causes of Panic Buying Following a Health Crisis. *International Journal of Environmental Research and Public Health*, *17*(10), 3513. https://doi.org/10.3390/ijerph17103513

Zhang, H., & Shaw, R. (2020). Identifying Research Trends and Gaps in the Context of COVID-19. *International Journal of Environmental Research and Public Health*, *17*(10), 3370. https://doi.org/10.3390/ijerph17103370